\documentclass[10pt]{article}

\usepackage[a4paper,textwidth=18cm,textheight=24cm,top=2.85cm, bottom=2.85cm, left=1.5cm, right=1.5cm]{geometry}

\usepackage{icdp2009}

\makeatletter
\long\def\@makecaption#1#2{%
\vskip\abovecaptionskip
\sbox\@tempboxa{#1. #2}%
\ifdim \wd\@tempboxa >\hsize
#1. #2\par
\else
\global \@minipagefalse
\hb@xt@\hsize{\box\@tempboxa\hfil}%
\fi
\vskip\belowcaptionskip}
\makeatother

\usepackage{lmodern}

\usepackage{graphicx}
\usepackage{times}

\begin{document}
\noindent

\bibliographystyle{plain}

\title{A multi-feature tracking algorithm enabling adaptation to context variations}

\authorname{Duc Phu CHAU, Fran\c{c}ois BREMOND, Monique THONNAT}

\authoraddr{Pulsar team, INRIA Sophia Antipolis - M\'editerran\'ee, France}

\maketitle

\keywords
Tracking algorithm, tracking features, Adaboost.
% seven videos in which six sequences belong to three public datasets: TRECVid \cite{trecvid}, ETISEO \cite{etiseo}, Caviar \cite{caviar} and the last one is provided by the Caretaker project. 

\abstract
We propose in this paper a tracking algorithm which is able to
adapt itself to different scene contexts. A feature pool is used to
compute the matching score between two detected objects. This feature pool includes 2D, 3D displacement distances,
2D sizes, color histogram, histogram of oriented gradient (HOG),
color covariance and dominant color. An offline learning process is proposed to
search for useful features and to estimate their weights for each context. In the
online tracking process, a temporal window is defined to
establish the links between the detected objects. This enables to find the object trajectories even if the objects are misdetected in some frames. A trajectory filter is proposed to remove noisy trajectories. Experimentation on different contexts is shown. The proposed tracker has been tested in videos belonging to three public datasets and to the Caretaker European project. The experimental results prove the effect of the proposed feature weight learning, and the robustness of the proposed tracker compared to some methods in the state of the art. The contributions of our approach over the state of the art trackers are:
(i) a robust tracking algorithm based on a feature pool, (ii) a
supervised learning scheme to learn feature weights for each
context, (iii) a new method to quantify the reliability of
HOG descriptor, (iv) a combination of color covariance and dominant color features
with spatial pyramid distance to manage the case of object occlusion.

%%%%%%%%% BODY TEXT
\section{Introduction}
Many approaches have been proposed to track mobile objects in a scene \cite{Yilmaz}. The problem is to have tracking algorithms which perform well in different scene conditions (e.g. different people density levels, different illumination conditions) and to be able to tune their parameters. The ideas of an automatic control for adapting an algorithm to the context variations  have already been studied \cite{monique, hall, prost}. In \cite{monique}, the authors have presented a framework which integrates knowledge and uses it to control image processing programs. However, the construction of a knowledge base requires a lot of time and data. Their study is restricted to static image processing (no video). In \cite{hall}, the author has presented an architecture for a self-adaptive perceptual system in which the ''auto-criticism`` stage plays the role of an online evaluation process. To do that, the system computes trajectory goodness score based on clusters of typical trajectories. Therefore, this method can be only applied for the scenes where mobile objects move on well defined paths. In \cite{prost}, the authors have presented a tracking framework which is able to control a set of different trackers to get the best possible performance. The approach is interesting but the authors do not describe how to evaluate online the tracking quality and the execution of three trackers in parallel is very expensive in terms of processing time.

In order to overcome these limitations, we propose a tracking algorithm that is able to adapt itself to
different contexts. The notion of context mentioned in this paper includes
a set of scene properties: density of mobile objects, frequence of
occlusion occurrences, illumination intensity, contrast level and the depth of the
scene. These properties have a strong effect on the tracking quality. In order to be able to track object movements in different contexts, we define firstly a feature pool in which each weighted feature combination
can help the system to outperform its performance in each context. However, the parameter configuration of these features (i.e.  determination of feature weight values) is a hard task because the user has to quantify correctly the
importance of each feature in the considered context. To
facilitate this task, we propose an offline learning algorithm based
on Adaboost \cite{adaboost} to compute feature weight values for each context. In this work, we have two assumptions. First, each video has a stable context. Second, for each context, there exists a training video set.

The paper is organized as follows: The next section presents the feature pool and explains how to use it to compute link similarity between the detected objects. Section 3 describes the offline learning process to tune the feature weights for each scene context. Section 4 shows in detail the different stages of the tracking process. The results of the experimentation and validation can be found in section 5. A conclusion as well as future work are given in the last section.

%------------------------------------------------------------------------
\section{Feature pool and link similarity}
%-------------------------------------------------------------------------
\subsection{Feature pool}
\label{secFeatures}
The principle of the proposed tracking algorithm is based on the coherence of mobile object features throughout time. In this paper, we define a set of 8 different features to compute a link similarity between two mobile objects $l$ and $m$ within a temporal window (see figure \ref{figGraph}).

\subsubsection{2D and 3D displacement distance similarity}
Depending on the object type (e.g. car, bicycle, walker), the object speed cannot exceed a fixed threshold. Let $D_{max}$ be the possible maximal 3D displacement of a mobile object for one frame in a video and $d$ be the 3D distance of two considered objects, we define a similarity $LS_1$ between these two objects using the 3D displacement distance feature as follows:
\begin{equation}
\label{eqDistance}
LS_1 = max (0,\ 1 - d/(D_{max} * n))
\end{equation}
\noindent where $n$ is the temporal difference (frame unity) of the two considered objects. 

Similarly, we also define a similarity $LS_2$ between two objects using displacement distance feature in the 2D image coordinate system.

%-------------------------------------------------------------------------
\subsubsection{2D shape ratio and area similarity}
\label{secShapeRatio}
Let $W_l$ and $H_l$ be the width and height
of the 2D bounding box of object $l$. The 2D shape ratio and area of this object are respectively defined as $W_l/H_l$ and $W_lH_l$. If no occlusions occur and mobile objects are well detected, shape ratio and area of a mobile object within a temporal window does not vary much even if the lighting and contrast conditions are not good. A similarity $LS_3$ between two 2D shape ratios of objects $l$ and $m$ is defined as follows:
\begin{equation}
\label{eqShapeRatio}
LS_3 = min (W_l/H_l,\ W_m/H_m)/max (W_l/H_l,\ W_m/H_m)
\end{equation}

Similarly, we also define the similarity $LS_4$ between two 2D areas of objects $l$ and $m$ as follows:
\begin{equation}
\label{eqArea}
\begin{small}LS_4 = min (W_lH_l,\ W_mH_m)/max (W_lH_l,\ W_mH_m) 
\end{small}\end{equation}

%-------------------------------------------------------------------------
\subsubsection{Color histogram similarity}
In this work, the color histogram of a mobile object is
defined as a normalized RGB color histogram of moving pixels inside its bounding box. We define a link similarity $LS_5$ between two objects $l$ and $m$ for color histogram feature as follows:
\begin{equation}
\label{eqColor}
LS_5 = \frac {\sum_{k = 1}^{ 3 \times K} min (H_l(k), H_m(k)) } {3}
\enspace
\end{equation}

\noindent where $K$ is a parameter representing the number of histogram bins for each color channel ($K = 1..256$), $H_l(k)$ and $H_m(k)$ are respectively the histogram values of object $l$, $m$ at bin $k$.

%-------------------------------------------------------------------------
\subsubsection{HOG similarity}
In case of occlusion, the system may fail to detect the full appearance of mobile objects. The above features are then unreliable. In order to address this issue, we propose to use the HOG descriptor to track locally interest points on mobile objects and to compute the trajectory of these points. The HOG similarity between two objects is defined as a value proportional to the number of pairs of tracked points belonging to both objects. In \cite{piotr}, the authors propose a method to track FAST points based on their HOG descriptors. However the authors do not compute the reliability level of the obtained point trajectories. In this work, we define a method to quantify the reliability of the trajectory of each interest point by considering the coherence of the Frame-to-Frame (F2F) distance, the direction and the HOG similarity of the points belonging to a same trajectory. We assume that the variation of these features follows a Gaussian distribution.

Let $(p_1, p_2, ..., p_i)$ be the trajectory of a point. Point $p_{i}$ is on the current tracked object and point $p_{i-1}$ is on an object previously detected. We define a coherence score $S_i^{dist}$ of F2F distance of point $p_i$ as follows:
\begin{equation}
S^{dist}_i = \frac{1}{\sqrt{2\pi\sigma_i^2}} e^{-\frac{(d_i - \mu_i)^2}{2\sigma_i^2}}
\end{equation}
\noindent where $d_i$ is the 2D distance between $p_i$ and $p_{i - 1}$, $\mu_i$ and $\sigma_i$ are respectively the mean and standard deviation of the F2F distance distribution formed by the set of points $(p_1, p_2, ..., p_i)$.

In the same way, we compute the direction coherence score $S^{dir}_i$ and the similarity coherence score $S^{desc}_i$ of each interest point. Finally for each interest point $p_i$ on the tracked object $l$, we define a coherence score $S_i^l$ as the mean value of these three coherence scores.

Let $P$ be the set of interest point pairs which trajectories pass through two considered objects $o_l$ and $o_m$; $S_i^l$ ($S_j^m$ respectively) be the coherence score of point $i$ ($j$ respectively) on object $l$ ($m$ respectively) belonging to set $P$. We define the similarity of HOG between these two objects as follows:
\begin{equation}
LS_6 = min (\frac{\sum_{i=1}^{|P|}S_i^l}{M_l}, \frac{\sum_{j=1}^{|P|}S_j^m}{M_m})
\end{equation}

\noindent where $M_l$ and $M_m$ are the total number of interest points detected on objects $l$ and $m$.

%-------------------------------------------------------------------------
\subsubsection{Color covariance similarity}
Color covariance is a very useful feature to characterize the appearance model of an image region. In particular, the color covariance matrix enables to compare regions of different sizes and is invariant to identical shifting of color values. This becomes an advantageous property when objects are tracked under varying illumination conditions. In \cite{sbak}, for a point $i$ in a given image region $R$, the authors define a covariance matrix $C_i$ corresponding to 11 descriptors:
$
\begin{footnotesize}
 \{x, y, R_{xy}, G_{xy}, B_{xy}, M_{xy}^R, O_{xy}^R, M_{xy}^G, O_{xy}^G, M_{xy}^B, O_{xy}^B\}
\end{footnotesize}
$
\noindent where ($x$, $y$) is pixel location, $R_{xy}, G_{xy},$ and $B_{xy}$ are RGB channel values, and $M$, $O$ correspond to gradient magnitude and orientation in each channel at position $(x, y)$.

We use the distance defined by \cite{forstner} to compare two covariance matrices:
\begin{equation}
 \rho (C_i, C_j) = \sqrt{\sum_{k=1}^F ln^2 \lambda_k(C_i, C_j)} 
\end{equation}
\noindent where $F$ is the number of considered image descriptors ($F = 11$ in this case), $\lambda_k(C_i, C_j)$ is the generalized eigenvalue of $C_i$ and $C_j$.

In order to take into account the spatial coherence of the color covariance distance and also to manage occlusion cases, we propose to use the spatial pyramid distance defined in \cite{Grauman}. The main idea is to divide the image region of a considered object by a set of sub-regions. For each level $i$ ($i \geq 0$), the considered region is divided by a set of $2^i$ x $2^i$ sub-regions. Then we compute the local color covariance distance for each pair of corresponding sub-regions. The computation of each sub-region pair helps to evaluate the spatial structure coherence between two considered objects. In the case of occlusions, the color covariance distance between two regions corresponding to occluded parts is very high. Therefore, we take only a half of the lowest color covariance distances (i.e. highest similarities) for each level to compute the final color covariance distance.

The similarity of this feature is defined as a function of the spatial pyramid distance:
\begin{equation}
 LS_7 = max (0,\ 1 - d_{cov}/D_{cov\_max})
\end{equation}
\noindent where $d_{cov}$ is the spatial pyramid distance of the color covariance between two considered objects, and $D_{cov\_max}$ is the maximum distance for two color covariance matrices to be considered as similar.

\subsubsection{Dominant color similarity}
Dominant color descriptor (DCD) has been proposed by MPEG-7 and is extensively used for image retrieval \cite{Yang}. This is a reliable color feature because it takes into account only important colors of the considered image region. DCD of an image region is defined as $F = \{\{c_i , p_i \},\ i = 1..A \}$ where $A$ is the total number of dominant colors in the considered image region, $c_i$ is a 3D RGB color vector, $p_i$ is its occurrence percentage, with $\sum_{i=1}^A p_i = 1$.

Let $F_1$ and $F_2$ be the DCDs of two image regions of considered objects. The dominant color distance between these two regions is defined using the similarity measure proposed in \cite{Yang}. Also, similar to the color covariance feature, in order to take into account the spatial coherence and occlusion cases, we propose to use the spatial pyramid distance for the dominant color feature. The similarity of this feature is defined in the function of the spatial pyramid distance as follows:
\begin{equation}
 LS_{8} = 1 - d_{DC}
\end{equation}

\noindent where $d_{DC}$ is the spatial pyramid distance of dominant colors between two considered objects.

%-------------------------------------------------------------------------
\subsection{Link similarity}
Using the eight features we have described above, a link similarity $LS (o_l,\ o_m)$ is defined as a weighted combination of feature similarities $LS_i$ between objects $o_l$ and $o_m$: 
\begin{equation}
\label{eqGS}
LS (o_l,\ o_m) =  \frac{\sum_{k=1}^{8} w_kLS_k} {\sum_{k=1}^{8} w_k}
\end{equation}
\noindent where $w_k$ is the feature weight (corresponding to its effectiveness), at least one weight is not null.

%-------------------------------------------------------------------------
\section{Learning feature weights}
\label{secLearning}
Each feature described above is effective for some particular context conditions. However, how can the user quantify correctly the feature significance for a given context? In order to address this issue, we propose in this paper an offline supervised learning process using the Adaboost algorithm \cite{adaboost}. First a weak classifier is defined per feature. Then a strong classifier which combines these eight weak classifiers (corresponding to the eight features) with their weights is learnt.
 
For each context, we select a learning video sequence representative of this context. First, for each object pair $(o_l, o_m)$ (called a training sample) in two consecutive frames, denoted $op_i$ ($i = 1..N$), we classify it into two classes \{+1, -1\}: $y_i = +1$ if the pair belongs to the same tracked object and $y_i = -1$ otherwise. For each feature $k\ (k = 1..8)$, we define a classification mechanism for a pair $op_i$ as follows:
\begin{equation}
\label{eqAdaboost}
h_k (op_i) =
\left\{
\begin{array}{l}
                   +1 \ \ if\ LS_k (o_l,\ o_m) \geq Th_1 \\
                   -1 \ \ otherwise
\end{array}
\right.
\end{equation}

\noindent where $LS_k (o_l,\ o_m)$ is the similarity score of feature $k$ (defined in section \ref{secFeatures})  between two objects $o_l$ and $o_m$, $Th_1$ is a predefined threshold representing the minimum feature similarity considered as similar.

The loss function for Adaboost algorithm at iteration $z$ for each feature $k$ is defined as:
\begin{equation}
\epsilon_k = \sum_{i=1}^N D_z(i)max(0, - y_ih_k(op_i))
\end{equation}
\noindent where $D_z(i)$ is the weight of the training sample $op_i$ at iteration $z$. At each iteration $z$, the goal is to find $k$ whose loss function $\epsilon_k$ is minimum. $h_k$ and $\epsilon_k$ (corresponding to value $k$ found) are denoted $h_z$ and $\epsilon_z$. The weight of this weak classifier denoted $\alpha_z$ is computed as follows:

\begin{equation}
\alpha_z = \frac{1}{2} ln \frac{1 - \epsilon_z}{\epsilon_z}
\end{equation}

We then update the weight of samples:
\begin{equation}
D_{z+1}(i) =
\left\{ 
\begin{array}{l}
	1/N \ , \ \ \ \ \ \ \ \ \ \ \ \ \ \ \ \ \ \ \ \ \ \ \ \ \ if\ z = 0 \\ \\
	\frac{D_z(i)exp(-\alpha_z y_i h_z(op_i) )}{A_z}, \ \ \ otherwise

\end{array}
\right.
\end{equation}

\noindent where $A_z$ is a normalization factor so that \begin{small}$\sum_i^N D_{z+1}(i)=1$\end{small}.

At the end of the Adaboost algorithm, the feature weights are determined for the learning context and allow to compute the link similarity defined in formula \ref{eqGS}.

%-------------------------------------------------------------------------
\section{The proposed tracking algorithm}
The proposed tracking algorithm needs a list of detected objects in a temporal window as input. The size of this temporal window (denoted $T_2$) is a parameter. The proposed tracker is composed of three stages. First, the system computes the link similarity between any two detected objects appearing in a given temporal window to establish possible links. Second, the trajectories that include a set of consecutive links resulting from the previous stage, are then computed as the system gets the highest possible total of global similarities (see section \ref{secTrajDet}). Finally, a filter is applied to remove noisy trajectories.

\subsection{Establishment of object links}
\label{secSetLink}
\begin{figure}[h]
\centering
\includegraphics[width=5.5cm]{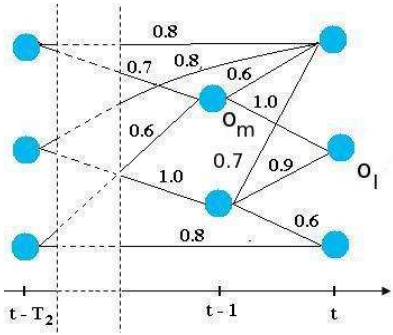}
\caption{The graph representing the established links of the detected objects in a temporal window of size $T_2$ frames.}
\label{figGraph}
\end{figure}

For each detected object pair in a given temporal window of size $T_2$, the system computes the link similarity (i.e. instantaneous similarity) defined in formula \ref{eqGS}. A temporal link is established between these two objects when their link similarity is greater or equal to $Th_1$ (presented in equation \ref{eqAdaboost}). At the end of this stage, we obtain a weighted graph whose vertices are the detected objects in the considered temporal window and whose edges are the temporally established links associated with the object similarities (see figure \ref{figGraph}).

%-------------------------------------------------------------------------
\subsection{Long term similarity}
\label{secLTSimil}
In this section, we study similarity score between an object $o_l$ detected at $t$ and the trajectory of $o_m$ detected previously, called long term similarity (to distinguish with the link similarity score between two objects). By assuming that the variations of the 2D area, shape ratio, color histogram, color covariance and dominant color features of a mobile object follow a Gaussian distribution, we can use the Gaussian probability density function (PDF) to compute this score. Also, longer the trajectory of $o_m$ is, more reliable this similarity is. Therefore, for each feature $k$ in these features, we define a long term similarity score between object $o_l$ and trajectory of $o_m$ as follows:
\begin{equation}
\label{eqLT}
LT_k(o_l,\ o_m) = \frac{1}{\sqrt{2\pi\sigma_m^2}} e^{-\frac{(s_l - \mu_m)^2}{2\sigma_m^2}} min (\frac{T}{Q}, 1)
\end{equation}

\noindent where $s_l$ is the value of feature $k$ for object $l$, $\mu_m$ and $\sigma_m$ are respectively mean and standard deviation values of feature $k$ of last $Q$-objects belonging to the trajectory of $o_m$ ($Q$ is a predefined parameter), $T$ is time length (number of frames) of $o_m$ trajectory. Thanks to the selection of the last $Q$-objects, the long term similarity can take into account the latest variations of the $o_m$ trajectory. 

For the left features (2D, 3D displacement distance and HOG), the long term similarity are set to the same values of link similarity.

%-------------------------------------------------------------------------
\subsection{Trajectory determination}
\label{secTrajDet}
The goal of this stage is to determine the trajectories of the mobile objects. For each detected object $o_l$ at instant $t$, we consider all its matched objects $o_m$ (i.e. objects with temporal established links) in previous frames that do not have yet official links (i.e. trajectories) to any objects detected at $t$. For such an object pair $(o_l,\ o_m)$, we define a global score $GS(o_l,\ o_m)$ as follows:
\begin{equation}
GS(o_l,\ o_m) = \frac{\sum_{k=1}^{8}w_k GS_k(o_l,\ o_m)}{\sum_{k=1}^{8}w_k}
\end{equation}

\noindent where $w_k$ is the weight of feature $k$ (resulting from learning phase, see section \ref{secLearning}), $GS_k(o_l,\ o_m)$ is the global score of feature $k$ between $o_l$ and $o_m$, defined as a function of link similarity and long term similarity of feature $k$:
\begin{equation}
GS_k(o_l,\ o_m) = (1 - \beta) LS_k(o_l,\ o_m)+\beta LT_k(o_l,\ o_m)
\end{equation}

\noindent where $LS_k(o_l,\ o_m)$ is the link similarity of feature $k$ between the two objects $o_l$ and $o_m$, $LT_k(o_l,\ o_m)$ is their long term similarity defined in section \ref{secLTSimil}, $\beta$ is the weight of long term similarity and is defined as follows:
\begin{equation}
\beta = min ( \frac{T}{Q}, Th_4  )
\end{equation}
\noindent where $T$, $Q$ are presented in section \ref{secLTSimil}, and $Th_4$ is the maximum expected weight for the long term similarity.

The object $o_m$ having the highest global similarity is considered as a temporal father of object $o_l$. After considering all objects at instant $t$, if more than one object get $o_m$ as a father, the pair $(o_l,\ o_m)$ which $GS(o_l,\ o_m)$ value is the highest will be kept and the link between this pair is official (i.e. become officially a trajectory segment). An object is no longer tracked if it cannot establish any official links in $T_2$ consecutive frames.

%-------------------------------------------------------------------------
\subsection{Trajectory filtering}
Noise usually appears when wrong detection or misclassification (e.g. due to low image quality) occurs. Hence a static object (e.g. a chair, a machine) or some image regions
(e.g. window shadow, merged objects) can be detected as a mobile object. However, such noise usually only appears in few frames or have no real motion. We thus use temporal and spatial filters to remove potential noises. A trajectory is considered as a noise if one of the two following conditions is satisfied:
$$
\begin{array}{lcl}
T & < & Th_5 \\
d_{max} & < & Th_6
\end{array}
$$
\noindent where $T$ is time length of the considered trajectory; $d_{max}$ is the maximum spatial length of this trajectory; $Th_5$, $Th_6$ are predefined thresholds.

%-------------------------------------------------------------------------
\section{Experimentation and Validation}
The objective of this experimentation is to prove the effect of feature weight learning, also to compare the performance of the proposed tracker with other trackers in the state of the art. To this end, in the first part, we test the proposed tracker with two complex videos (many moving people, high occlusion occurrence frequency) which are respectively provided by the Caretaker European project\footnote{http://cordis.europa.eu/ist/kct/caretaker\_synopsis.htm} and the TRECVid dataset \cite{trecvid}. These two videos are tested in both cases: without and with the feature weight learning. In the second part, five videos belonging to two public datasets ETISEO\footnote{http://www-sop.inria.fr/orion/ETISEO/} and Caviar\footnote{http://homepages.inf.ed.ac.uk/rbf/CAVIARDATA1/} are experimented, and the tracking result (with the feature learning) is compared with some other approaches in the state of the art.

In order to evaluate the tracking performance, we use the three tracking evaluation metrics defined in the ETISEO project \cite{atnghiem}. The first tracking evaluation metric $M_1$ measures the percentage of time during which a reference object (ground truth data) is correctly tracked. The second metric $M_2$ computes throughout time how many tracked objects are associated with one reference object. The third metric $M_3$ computes the number of reference object IDs per tracked object. These metrics must be used together to obtain a complete performance evaluation. Therefore, we also define a tracking metric $\overline{M}$ taking the average value of these three tracking metrics. The four metric values are defined in the interval [0, 1]. The higher the metric value is, the better the tracking algorithm performance gets.

In this experimentation, we use the people detection algorithm based on the HOG descriptor of the OpenCV library. So we focus the experimentation on the sequences containing people movements. However the principle of the proposed tracking algorithm is not dependent on the tracked object type. For learning feature weights, we use video sequences that are different from the tested videos but which have a similar context.

\begin{figure*}[t]
\centering
\includegraphics[width=17cm]{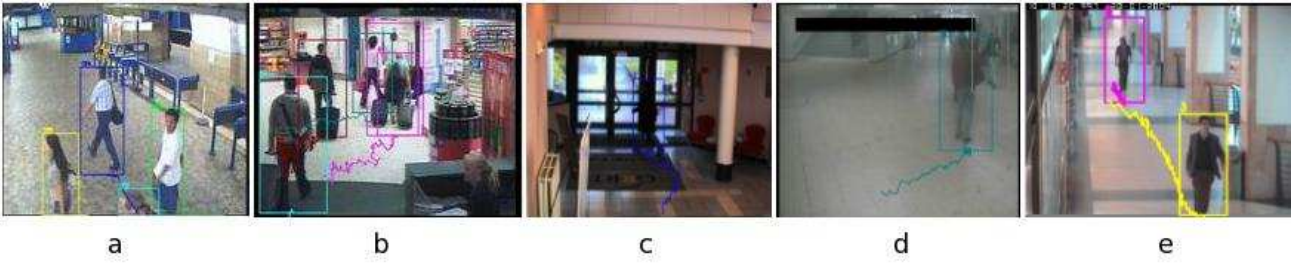}
\begin{small}
\caption{Illustration of five tested videos: a. Caretaker b. Trecvid c. ETI-VS1-BE-8-C4 d. ETI-VS1-MO-7-C1 e. Caviar}
\end{small}
\label{fig_videos}  
\label{figETI}
\end{figure*}

The first tested video (provided by the Caretaker project) depicts people moving in a
subway station. The
frame rate of this sequence is $5$ $fps$ ($frames/second$) and the length is 5 min (see image \ref{fig_videos}a). We have learnt feature weights on a sequence of 2000 frames. The learning algorithm selects $w_5 = 0.5$ (color histogram feature) and $w_6 = 0.5$ (HOG feature).

The second tested sequence (belonging to the TRECVid dataset) depicts the movements of people in an airport (see image \ref{fig_videos}b). It contains 5000 frames and lasts 3 min 20 sec. We have learnt feature weights on a sequence of 5000 frames. The learning algorithm selects $w_1 = 0.24$ (3D distance displacement), $w_4 = 1$ (2D area) and $w_5 = 0.76$ (color histogram).

Table \ref{tab_car_trec} presents the tracking results in two cases: without and with feature weight learning. We can find that with the proposed learning scheme, the tracker performance increases in both tested videos. Also, the processing time of the tracker also decreases significantly because many features are not used.
\begin{table}[b]
   \begin{center}
	\begin{tabular}{|p{2.05 cm}|p{0.4 cm}|p{0.4 cm}|p{0.4 cm}|p{0.4 cm}|p{0.4 cm}|p{0.4 cm}|p{0.4 cm}|p{0.4 cm}|}
	
		\hline
			  & \multicolumn{4}{|c|}{Without learning} & \multicolumn{4}{|c|}{With learning} \\
 		\hline
			 &  $M_1$ & $M_2$ & $M_3$ & $\overline{M}$ 	&  $M_1$ & $M_2$ & $M_3$ & $\overline{M}$ \\
		\hline
\begin{small}Caretaker video\end{small}&0.62& 0.16  & 0.99  &0.59&  0.47  & 0.83  & 0.80  & 0.70		 \\
		\hline

\begin{small}TRECVid video\end{small}&  0.60  & 0.82  & 0.90  &0.77&  0.70	 & 0.93  & 0.84  & 0.82 \\
		\hline
	\end{tabular}
	\end{center}
\caption{\label{tab_car_trec}Summary of tracking results in both cases: without and with feature weight learning.}
\end{table}

The two following tested videos belong to ETISEO dataset. The first tested ETISEO video shows a building entrance, denoted ETI-VS1-BE-18-C4. It contains 1108 frames and frame rate is 25 $fps$. In this sequence, there is only one person moving (see image \ref{fig_videos}c). We have learnt feature weights on a sequence of 950 frames. The learning algorithm has selected the 3D displacement distance feature as the unique feature for tracking in this context. The result of the learning phase is reasonable since there is only one moving person.

The second tested ETISEO video shows an underground station denoted ETI-VS1-MO-7-C1 with occlusions. The difficulty of this sequence consists in the low contrast and bad illumination. The scene depth is quite important (see image \ref{fig_videos}d). This video sequence contains 2282 frames and frame rate is 25 $fps$. We have learnt feature weights on a sequence of 500 frames. The color covariance feature is selected as the unique feature for tracking in this context. It is a good solution because the dominant color and HOG feature do not seem to be effective due to bad illumination. Also, the size and displacement distance features are not reliable because their measurements do not seem to be discriminative for far away moving people from the camera.

In these two experiments, tracker results from seven different teams (denoted by numbers) in ETISEO have been presented: 1, 8, 11, 12, 17, 22, 23. Because names of these teams are hidden, we cannot determine their tracking approaches. Table \ref{tabRes} presents performance results of the considered trackers. The tracking evaluation metrics of the proposed tracker get the highest values in most cases compared to other teams.
\begin{table}[]
\begin{footnotesize}
\begin{center}
		\begin{tabular}{|p{0.5 cm}|p{0.5 cm}|p{0.5 cm}|p{0.5 cm}|p{0.5 cm}|p{0.5 cm}|p{0.5 cm}|p{0.5 cm}|p{0.5 cm}|}

		\hline

			&\multicolumn{2}{|c|}{Our tracker} & \multicolumn{2}{|c|}{Team 1} & \multicolumn{2}{|c|}{Team 8}&\multicolumn{2}{|c|}{Team 11}\\
		\cline{2-9}

				&BE		&MO 		&BE 	&MO		&BE	&MO 		&BE 		&MO 	\\

		\hline
			$M_1$	&0.50		&0.79	 	&0.48 	&0.77		&0.49	&0.58		&\textbf{0.56} &0.75	\\

			$M_2$	&\textbf{1.00}	&\textbf{1.00}	&0.80 	&0.78		&0.80 	&0.39		&0.71		&0.61	\\

			$M_3$ 	&\textbf{1.00}	&\textbf{1.00}	&0.83 	&\textbf{1.00}	&0.77 	&\textbf{1.00}	&0.77		&0.75	\\

		$\overline{M}$	&\textbf{0.83}	&\textbf{0.93} 	&0.70	&0.85		&0.69	&0.66		&0.68		&0.70	\\
		\hline

			&\multicolumn{2}{|c|}{Team 12} & \multicolumn{2}{|c|}{Team 17} & \multicolumn{2}{|c|}{Team 22}&\multicolumn{2}{|c|}{Team 23}\\
		\cline{2-9}

				&BE		&MO 		&BE 	&MO		&BE	&MO 		&BE 		&MO 	\\

		\hline
			$M_1$	&0.19		&0.58	 	&0.17	&\textbf{0.80}	&0.26	&0.78		&0.05	&0.05	\\

			$M_2$	&\textbf{1.00}	&0.39		&0.61	&0.57		&0.35	&0.36		&0.46	&0.61	\\

			$M_3$ 	&0.33		&\textbf{1.00}	&0.80	&0.57		&0.33	&0.54		&0.39	&0.42	\\

		$\overline{M}$	&0.51		&0.66 		&0.53	&0.65		&0.31	&0.56		&0.30	&0.36	\\
		\hline

		\end{tabular}
		\caption{\label{tabRes}Summary of tracking results for two ETISEO videos. BE denotes ETI-VS1-BE-18-C4 sequence, MO denotes ETI-VS1-MO-7-C1 sequence. The highest values are printed bold.}
\end{center}
\end{footnotesize}
\end{table}

The last three tested videos belong to the Caviar dataset (see image \ref{fig_videos}e). In this dataset, we have selected the same sequences experimented in \cite{snidaro} to be able to compare each other: OneStopEnter2cor, OneStopMoveNoEnter1cor and OneStopMoveNoEnter2cor. In these three sequences, there are 9 persons walking in a corridor. The proposed approach can track all of them. However there are three noisy trajectories in the last sequence because of wrong detection occurred in a long period. Table \ref{tab_caviar} presents the result summary for these videos. TP (True Positive) refers to the number of correct tracked trajectories. FN (False Negative) is the number of lost trajectories. FP (False Positive) represents the number of noisy trajectories. Compared to \cite{snidaro}, our proposed tracker have better values in all of these three indexes.
\begin{table}[t]
\begin{small}
	\begin{center}
% 		\begin{tabular}{ |p{2.3 cm}|p{1.9 cm}|p{0.4 cm}|p{0.4 cm}| p{0.4 cm}|}
		\begin{tabular}{ |c|c|c|c|c|}
 		\hline
	 		& \# trajectories & TP & FN & FP \\
		\hline
	Proposed tracker&	9	&	9	& 0  		 & 3 \\
		\hline
Approach of \cite{snidaro} 	&  9 & 	8	&1		  & 7	\\
		\hline

		\end{tabular}
		\caption{\label{tab_caviar}Summary of tracking results for three Caviar videos}

	\end{center}
\end{small}
\end{table}
%-------------------------------------------------------------------------
\section{Conclusion and Future work}
We have presented in this paper an approach which combines a large set of appearance features and learn tracking parameters. The quantification of HOG descriptor reliability and the combination of color covariance, dominant color with spatial pyramid distance help to increase the robustness of the tracker for managing occlusion cases. The learning of feature significances for different video contexts also helps the tracking algorithm to adapt itself to the context variation problem. The experimentation proves the effect of the feature weight learning, also the robustness of the proposed tracker compared to some other approaches in the state of the art. We propose in future work an automatic context detection to increase the auto-control capacity of the system and to remove the two assumptions given in this paper (presented in section 1).

\section*{Acknowledgement}
\noindent This work is supported by The PACA region, The General Council of Alpes Maritimes province, France as well as The ViCoMo, Vanaheim and Support projects.

\bibliography{icdp2009}

\end{document}